\documentclass[12pt]{article}

\usepackage{amsmath}
\usepackage{amsfonts}

\newtheorem{theorem}{Theorem}

\newtheorem{definition}[theorem]{Definition}

\newtheorem{proposition}[theorem]{Proposition}

\begin{document}

\title{Classification by ensembles of neural networks}

\author{S.V. Kozyrev\footnote{Steklov Mathematical Institute}}

\maketitle

\begin{abstract}
We introduce a new procedure for training of artificial neural networks by using the approximation of an objective function by arithmetic mean of an ensemble of selected randomly generated neural networks, and apply this procedure to the classification (or pattern recognition) problem. This approach differs from the standard one based on the optimization theory. In particular, any neural network from the mentioned ensemble may not be an approximation of the objective function.
\end{abstract}

\section{Introduction}

The standard approach to artificial neural networks is based on the optimization theory, cf. for example  \cite{NikolenkoTulupiev}. Artificial neural network is a composition of neurons (see the next section), which depends on the set of real parameters (weights of the neural network). In the pattern recognition (or classification) problem a neural network is considered as an approximation of an objective function (characteristic function of an objective, or target, set) at the training set. In order to find this approximation the optimization problem (minimization of the norm of the difference between the objective function and the neural network at the training set) at the space of parameters of the neural network is studied.

In the present paper we introduce a new approach to training of (ensembles of) artificial neural networks. In this approach instead of optimization of the parameters of a single neural network we consider an ensemble of selected neural networks with randomly chosen parameters. A neural network is selected if this network has a sufficiently small number of errors at the training set. We introduce the {\it averaged neural network} which in the simplest case is an arithmetic mean of selected neural networks. We show that the averaged neural network can be considered as some kind of approximation of the objective function.

Using the introduced in the present paper approach we are able to avoid the two general problems of theory of neural networks: the problem of global optimization at a complex landscape and the problem of overfitting.

The exposition of the present paper is as follows.

In Section 2 we introduce the necessary notations and discuss the standard approach to training of neural networks based on the approximation theory.

In Section 3 we introduce some procedure based on selection and averaging as a method of training of (ensembles of) neural networks.

In Section 4 we discuss relation of the construction of the present paper and theory of biological evolution.

\section{Training of neural networks and approximation theory}

Let us recall some definitions of theory of artificial neural networks.

A neuron (or single layer perceptron) is a function of $N$ real variables of the form
\begin{equation}\label{perceptron}
f(x_1,\dots,x_N)={\rm sgn}\left(\sum_{i=1}^Nw_ix_i-\theta\right).
\end{equation}
Here $x_i$ are real variables, $(x_1,\dots,x_N)$ takes values in some domain $U\subset \mathbb{R}^N$, $w_i$ are real parameters (weights of the neuron), $\theta$ is the threshold of activation of the neuron, the function ${\rm sgn}(x)=1$ for $x\ge 0$ and is equal to zero for $x<0$.

We also consider the smoothed variant of the above neuron for which instead of the function ${\rm sgn}$ we use the smooth monotonous increasing function ${\rm sgm}$ which varies from zero to unity. In particular we consider the neuron of the form
\begin{equation}\label{sigmoid}
f(x_1,\dots,x_N)={\rm sgm}\left(\sum_{i=1}^Nw_ix_i-\theta\right),\qquad {\rm sgm}(x)={1\over 1+e^{-x}}.
\end{equation}

A neural network is a composition of the above neurons.

\bigskip

\noindent{\bf Example}\quad Let us consider a double layer neural network with the neurons of the form (\ref{perceptron})
\begin{equation}\label{double_layer}
f(x_1,\dots,x_N)={\rm sgn}\left(\sum_{i=1}^Kw_iy_i-\theta\right),\qquad y_i={\rm sgn}\left(\sum_{j=1}^Nw_{ij}x_j-\theta_i\right).
\end{equation}

This function is equal to one at some final family of the sets $\{y_i\}$, $y_i=0,1$, $i=1,\dots,K$. For any set $\{y_i\}$ from this family the corresponding $x_i$ have to satisfy the system of inequalities
$$
\sum_{j=1}^Nw_{ij}x_j\ge\theta_i,\quad y_i=1, \qquad \sum_{j=1}^Nw_{ij}x_j<\theta_i,\quad y_i=0,\qquad i=1,\dots,K,
$$
i.e. to belong to the intersection of $K$ half--spaces of the dimension $N$.

Therefore the above neural network (\ref{double_layer}) is equal to characteristic function of a finite union of intersections of half--spaces.

\bigskip

Let us discuss the classification problem for neural networks. Let the domain $U$ of $f$ be a union of the two parts --- the objective set  $T$ and the complement $U\backslash T$ of this set.

The classification problem is as follows: to build an approximation of the objective function (the characteristic function $\chi_T$ of the objective set $T$) by a neural network $f$, i.e. by a composition of neurons of the form (\ref{perceptron}) or (\ref{sigmoid}).

Let us recall some definitions of the approximation theory. Let $V$ be a normed linear space and $M\subset V$ is some subset. An approximation of $g\in V$ by $f\in M$ have to satisfy
$$
\|g-f\|={\rm inf}_{f'\in M}\|g-f'\|,
$$
where $\|\cdot\|$ is the norm in $V$ (i.e. we find the element of $M$ nearest to $g\in V$).

For neural networks the space $V$ is a subset of $L^2(U)$ (with the corresponding norm), $M$ is the set of neural networks with the fixed architecture (i.e. the form of the composition of neurons is fixed but the weights $w_i$ and activation thresholds $\theta$ for neurons are the parameters of $f\in M$).

Our aim is to approximate the objective function $\chi_T$ by $f\in M$. The problem is that we do not know the exact form of the objective set $T$. Instead we have the training set $X$ --- the finite family of elements $x\in U$ for which we know, which of the elements $x\in X$ belongs to the objective set $T$ and which $x\in X$ does not belong to $T$.

This implies the following definition: the solution of the classification problem is the neural network $f$ with the parameters $w_i$, $\theta$ for which the rms (root mean square) deviation of the neural network from the objective function at the training set is minimal. Therefore the classification problem takes the form of some global optimization problem in the set of parameters of neural networks with the given architecture.

For the investigation of this optimization problem neural networks with smooth neurons of the form (\ref{sigmoid}) are applied (because optimization methods such as steepest descent are used). The other approaches to optimization are Monte Carlo method, simulated annealing and other methods.

There are the two main problems with training of neural network in the framework of opti\-mization. First, global nonlinear optimization is a computationally hard problem for the case of multiple local minima.

Second, there is a problem of overfitting --- our neural network may approximate not the objective function but the particular choice of the training set.

In the present paper we propose the alternative approach to classification with the help of neural networks which in some sense is free of the above problems. In this approach instead of finding the global minimum of the optimization problem we will take into account the contributions from the ensemble of local minima.

\section{Selection and averaging of neural networks}

Let us consider the set of neurons of the form (\ref{perceptron}) where $w_i$ and  $\theta$ are independent real random variables with some distributions. For simplicity we consider random variables with equal distributions. Using this ensemble of neurons we build the ensemble of neural networks \linebreak $\{f[w,\theta](x_1,\dots,x_N)\}$ as ensemble of compositions of neurons with independent random parameters (i.e. the form $f$ of the composition is given and the parameters $w$, $\theta$ are chosen independently for any neuron).

Let $X$ be the training set. Using the described above ensemble of neural networks we choose randomly a set of neural networks from this ensemble (corresponding to some choices of the random parameters $w$, $\theta$) in the following way. All neural networks from this set take the required values on the training set (i.e. these neural networks take values equal to one for $x\in X$ from the objective set and take values equal to zero for $x\in X$ from the complement to the objective set). This choice of neural networks corresponds to selection of neural networks at the training set.

The ensemble of selected neural networks $f[w,\theta](x_1,\dots,x_N)$ can be described as follows. The distribution functions for the parameters $w$, $\theta$ of a neural network from the initial ensemble are multiplied by the characteristic function of the set of parameters for which the corresponding neural network will take the required values at the training set. After this procedure the joint distribution function of the parameters $w$, $\theta$ have to be normalized (since the multiplication by the mentioned characteristic function breaks the normalization condition). The joint distribution function of the parameters $w$, $\theta$ for neural network from the initial ensemble is equal to the product of distributions of all the parameters (since the parameters are chosen independently). After the selection procedure the parameters $w$, $\theta$ of a neural network are no longer independent.

\begin{definition}{\sl Let us consider the finite set  $f[w^a,\theta^a]$, $a=1,\dots,n$ of independent selected neural networks.
We introduce the averaged neural network $\langle f \rangle$ as the limit of arithmetic means of selected neural networks
\begin{equation}\label{replica}
\langle f \rangle (x_1,\dots,x_N)=\lim_{n\to\infty}\langle f \rangle_n (x_1,\dots,x_N)=\lim_{n\to\infty}{1\over n}\sum_{a=1}^nf[w^a,\theta^a](x_1,\dots,x_N).
\end{equation}
}
\end{definition}

Therefore the averaged neural network takes the required values at the training set. Here the parameters $w^a$, $\theta^a$ for different selected neural networks are independent (they are not necessarily independent for a fixed network).

Selected neural networks may have different realizations (which we enumerate by $a$) but as random functions selected neural networks are equal. Thus the following expectation of a selected random network (with respect to the described above distribution of the parameters $w$, $\theta$) will not depend on $a$
$$
E\left(f[w^a,\theta^a](x_1,\dots,x_N)\right).
$$

The main statement of the present paper is that in the limit of large $n$ the averaged neural network $\langle f \rangle_n$ will be a solution of the classification problem, i.e. it will converge in probability to non--random function $\langle f \rangle$ which in some sense can be considered as an approximation of the objective function (the characteristic function of the objective set) for the classification problem under consideration.

\begin{proposition}
{\sl Let the expectation and the dispersion of the random function $f[w,\theta](x_1,\dots,x_N)$ exist.
Then in the limit $n\to\infty$  the random function $\langle f \rangle_n$ given by (\ref{replica}) converges in probability pointwise to the non--random function
\begin{equation}\label{averaged}
\langle f \rangle(x_1,\dots,x_N)=E\left(f[w,\theta](x_1,\dots,x_N)\right).
\end{equation}

}
\end{proposition}

\noindent {\it Proof}\qquad
The proof is by the law of large numbers. Random functions $f[w^a,\theta^a]$ from the ensemble of selected neural networks are independent for different $a$ and take values 0 and 1. The dispersions of these random functions coincide (for fixed arguments $x_1,\dots,x_N$), therefore we can apply the law of large numbers which proves the existence of the limit in (\ref{replica}) and (\ref{averaged}). $\square$

\bigskip

Here the approximation of the objective function by the function $\langle f \rangle$ is not understood in the sense of the approximation theory as in the previous section (where the approximation of the objective function is the closest function from the family of functions of the given form).

By the construction the averaged neural network $\langle f \rangle$ takes values 1 and 0 at the elements of the training set which belong to the objective set and its complement correspondingly.

At the element $(x_1,\dots,x_N)\in U$ of the domain $U$ of the neural networks under consideration which does not belong to the training set $X$ values of a part of the summands in (\ref{replica}) will be equal to one and values of another part will be equal to zero. Therefore for such $(x_1,\dots,x_N)$ the averaged neural network (\ref{replica}) will take some value from $[0,1]$.

\bigskip

\noindent{\bf Example}\quad
Let us consider the double layer neural network
$$
f(x_1,\dots,x_N)={\rm sgn}\left(\sum_{i=1}^Kw_iy_i-\theta\right),\qquad y_i={\rm sgn}\left(\sum_{j=1}^Nw_{ij}x_j-\theta_i\right),
$$
where all the weights $w_i$, $w_{ij}$, $\theta$, $\theta_i$ are independent random variables. By the definition of selected neural network we choose randomly the family of parameters $w_i^a$, $w_{ij}^a$, $\theta^a$, $\theta_i^a$, $a=1,\dots,n$ for which the corresponding neural network $f$ will take the required values at the training set $X$. Therefore the averaged neural network
$$
\langle f \rangle (x_1,\dots,x_N)={1\over n}\sum_{a=1}^n{\rm sgn}\left(\sum_{i=1}^Kw_i^ay_i-\theta^a\right),\qquad y_i={\rm sgn}\left(\sum_{j=1}^Nw_{ij}^ax_j-\theta_i^a\right)
$$
will also take the required values at the training set.

As we discussed for the example at the previous section, any of the summands $f[w^a,\theta^a]$ (double layer neural networks) in the expression above is equal to the characteristic function of a finite union of intersections of $K$ half--spaces. Any of these summands can be far (in the sense of approximation theory) from the characteristic function of the objective set. In particular, it is possible that the summand $f[w^a,\theta^a]$ is a characteristic function of some polyhedron, and some part of this polyhedron lies in between the points of the training set which do not belong to the objective set. In this situation $f[w^a,\theta^a](x_1,\dots,x_N)$ will be equal to one for $(x_1,\dots,x_N)$ from this part of the mentioned polyhedron but it is natural to expect that $(x_1,\dots,x_N)$ does not belong to the objective set.

In summation over the ensemble of selected neural network we may have many such cases but any of these cases (for the particular $(x_1,\dots,x_N)$ and $f[w^a,\theta^a]$) has low probability since the random parameters $w^a$, $\theta^a$ for the different $a$ are independent. Therefore in summation in (\ref{replica}) the corresponding contributions will be small because of the normalization ${1\over n}$. Thus the averaged neural network will give a better approximation of the objective function (characteristic function of the objective set) in comparison to the summands in (\ref{replica}).

\bigskip

The summands $f[w^a,\theta^a]$ in (\ref{replica}) which give the required values at the elements of the training set in some approximation correspond to local minima of the root mean square (rms) deviation of the neural network $f[w,\theta]$ from the objective function. Therefore in (\ref{replica}) we sum over the local minima of the rms deviation instead of looking for the global minimum as in the optimization theory.

Therefore the computational problem of finding of the global minimum in our approach is exchanged to the problem of finding of an ensemble of local minima. We are interested in simplification of this problem. Also it is important to make the definition of the averaged neural network more robust to errors in the training set. We consider the following generalization of the averaged neural network.

\bigskip

\noindent{\bf Generalization of the definition of averaged neural network for the case with errors.}\quad
Let us consider a more general ensemble of independent neural networks $\{f[w^a,\theta^a]\}$, $a=1,\dots$. Neural networks from this ensemble belong to the initial ensemble of neural networks (without selection) i.e. these networks may make errors when applied to the elements of the training set $X$ (may be equal to one for $x\in X$ which lies outside the objective set or may be equal to zero for $x\in X$ which lies inside the objective set). Let the neural network $f[w^a,\theta^a]$ possesses $m \left(f[w^a,\theta^a]\right)$ errors at the training set $X$.

We introduce the averaged neural network as the $n\to\infty$ limit of finite linear combinations of independent neural networks
\begin{equation}\label{replica1}
\langle f \rangle (x_1,\dots,x_N)=\lim_{n\to\infty}\left(\sum_{a=1}^n e^{-\beta m \left(f[w^a,\theta^a]\right)}\right)^{-1}\sum_{a=1}^n e^{-\beta m \left(f[w^a,\theta^a]\right)}f[w^a,\theta^a](x_1,\dots,x_N).
\end{equation}
In the expression above the averaged neural network is a result of averaging over the Gibbs ensemble of independent random neural networks with the inverse temperature $\beta>0$. The energy of the $a$-th neural network is equal to the number $m (f[w^a,\theta^a])$ of errors of this network at the training set $X$.

In the limit $n\to \infty$, by the law of large numbers, expression (\ref{replica1}) converges in probability to the Gibbs average
$$
\langle f \rangle (x_1,\dots,x_N)=\left( E\left( e^{-\beta m \left(f[w,\theta]\right)}\right)\right)^{-1} E\left( e^{-\beta m \left(f[w,\theta]\right)}f[w,\theta](x_1,\dots,x_N)\right).
$$
Here $E$ is the expectation with respect to the initial ensemble of random neural networks.

The neural network (\ref{replica1}) is equal to one at the elements of the training set $X$ from the objective set which are correctly (without errors) recognized by all the elements $\{f[w^a,\theta^a]\}$ of the ensemble of random neural network (correspondingly, is equal to zero for correctly recognized elements of the training set from the complement to the objective set).

Since we allowed errors the ensemble $\{f[w^a,\theta^a]\}$ contains elements which are easier to generate in comparison to the case without errors considered earlier (\ref{replica}).
In the limit of zero temperature $\beta\to\infty$ the expression (\ref{replica1}) tends to the averaged neural network without errors (\ref{replica}). We have expressed the selection procedure with the help of averaging over the Gibbs ensemble.

\bigskip

\noindent{\bf Classification by ensembles of neural networks with different architectures.}\quad
One of the advantages of the approach proposed in the present paper  is the possibility to mix in the ensembles under consideration neural networks with different architectures, i.e. neural networks which are the different compositions of neurons of the form (\ref{perceptron}).

Let us consider the ensemble of neural networks containing neural networks with different architectures $f[w,\theta](x_1,\dots,x_N)$ (these networks will have the same domain, in particular will depend on the same number $N$ of variables, but the form of neural networks as compositions of neurons and the number of parameters $w$, $\theta$ may be different for different networks from the ensemble). Neural networks with the fixed architecture, as earlier, are generated randomly with the independent parameters $w$, $\theta$.

Let us introduce the generalization of the averaged neural network (\ref{replica1}) of the form
\begin{equation}\label{replica2}
\langle f \rangle (x_1,\dots,x_N)=\lim_{n\to\infty}{\sum_{a=1}^n  e^{-\beta \left( m \left(f^{c(a)}[w^a,\theta^a]\right) +k(c(a))\right)}f^{c(a)}[w^a,\theta^a](x_1,\dots,x_N)\over
\sum_{a=1}^n e^{-\beta\left( m \left(f^{c(a)}[w^a,\theta^a]\right) +k(c(a))\right)}}.
\end{equation}
Here the index $c$ enumerates the different architectures of neural networks $f^c$ (this index takes a finite number of values), $c(a)$ is the random architecture of the $a$-th randomly chosen neural network, the function $k(c)$ of complexity of the network takes positive values and increases sufficiently fast with the increasing of complexity of the neural network $f^c$ (in particular one can take $k(c)$ to be equal to the number of neurons in the network), the other notations have the same meaning as in (\ref{replica1}).

In the limit $n\to\infty$ expression (\ref{replica2}) will converge in probability to the Gibbs average
$$
\langle f \rangle (x_1,\dots,x_N)=\left( \sum_{c}e^{-k(c)} E\left( e^{-\beta m \left(f^{c}[w,\theta]\right) } \right) \right)^{-1}
\sum_{c}e^{-k(c)} E\left( e^{-\beta \left( m \left(f^{c}[w,\theta]\right)\right)}f^c[w,\theta](x_1,\dots,x_N)\right).
$$

For low temperature (large $\beta$) the main contribution to expression (\ref{replica2}) comes from neural networks which have sufficiently simple architectures and are able to solve the classification problem (i.e. to give considerable number of enumerated by the index $a$ contributions with small number of errors to expression (\ref{replica2})). Since (\ref{replica2}) contains contributions from neural networks with the different (in particular simple) architectures this will help to reduce the problem of overfitting of neural network --- optimization of a neural network of unnecessarily complicated architecture for the particular form of the training set which may cause errors for a different training set with the same objective function.

\section{Discussion}

The standard approach to the classification (or pattern recognition) problem with neural networks is as follows: we choose the architecture of the neural network and then find the parameters of the network which give a better approximation of the objective function, i.e. solve the optimization problem.

In the present paper we propose the alternative approach: instead of optimizing the particular neural network we consider the Gibbs ensemble of neural networks with different architectures and energy equal to the sum of the number of errors of the neural network at the training set and some increasing function of complexity of the neural network. Then for sufficiently low temperatures the Gibbs average over the ensemble of neural networks will give the solution of the classification problem.

Let us discuss the following analogue with the theory of biological evolution. In accordance with the modern approach in evolution theory, so called {\it postmodern synthesis} \cite{Koonin}, biological systems are ensembles of replicators (in particular genes). In \cite{Koonin} it is stressed that it is necessary to consider genomes from the point of view of statistical physics applied to genomic sequences.

As a development of this approach we propose to take into account the computational aspect of genomic sequences i.e. to consider genomes as ensembles of some simple algorithms. Any of these algorithms is a replicator (for example a gene). The key question in this approach will be the description of gene regulation as interaction of algorithms in the ensemble.

Why some ensemble of algorithms can function as a single algorithm? In particular, it is interesting to construct a simplest example of such an ensemble of algorithms which solves some problem.

The second question, why biological evolution is possible, i.e. why selection and other manipulations with statistical ensembles can generate sufficiently complex algorithms starting from an ensemble of elementary algorithms?

In the present paper we have constructed the ensemble of neural networks which solves the classification problem. Replication and mutation of the set of neural networks of different architectures were used, and the selection procedure with the help of the Gibbs ensemble described above was applied.

Let us note that in the standard approach to selection in evolution theory selection is considered as optimization procedure --- one has to select elements with higher fitness.
The main point of the approach of the present paper is the classification by an ensemble which contains neural networks with sufficiently different properties.

The introduced in this paper method can be applied to the description of evolution of a genome as an ensemble of algorithms. This approach will be some version of the theory of group selection applied not to population of individuals but to genome as ensemble of replicators.

The approach to genomes as probabilistic algorithms, in particular modeling of gene duplication by some analogue of replica procedure analogous to applied in theory of spin glasses \cite{MPV} was proposed in \cite{replica_algorithm}.

\bigskip

\noindent{\bf Acknowledgments}\qquad The author gratefully
acknowledges being partially supported by the grants of
the Russian Foundation for Basic Research
RFBR 11-01-00828-a and 11-01-12114-ofi-m-2011, by the grant of the President of Russian
Federation for the support of scientific schools NSh-2928.2012.1, by the DFG project AL 214/40-1 and
by the Program of the Department of Mathematics of the Russian
Academy of Science ''Modern problems of theoretical mathematics''.

\end{document}